# Comparison Training for Computer Chinese Chess

Wen-Jie Tseng[1], Jr-Chang Chen[2], I-Chen Wu[1], *Senior Member, IEEE*, Tinghan Wei[1]

*Abstract* —This paper describes the application of comparison training (CT) for automatic feature weight tuning, with the final objective of improving the evaluation functions used in Chinese chess programs. First, we propose an n-tuple network to extract features, since n-tuple networks require very little expert knowledge through its large numbers of features, while simultaneously allowing easy access. Second, we propose a novel evaluation method that incorporates tapered eval into CT. Experiments show that with the same features and the same Chinese chess program, the automatically tuned comparison training feature weights achieved a win rate of 86.58% against the weights that were hand-tuned. The above trained version was then improved by adding additional features, most importantly n-tuple features. This improved version achieved a win rate of 81.65% against the trained version without additional features.

*Index Terms*—comparison training, n-tuple networks, machine learning, Chinese chess

## I. Introduction

CHINESE chess is one of the most popular board games worldwide, with an estimated player base of hundreds of millions of people [28]. It is a two-player zero-sum game. The state-space complexity and game-tree complexity of Chinese chess are $10^{48}$ and $10^{150}$ respectively, which are between those of shogi and chess [1][11][12].

Most strong Chinese chess programs, such as SHIGA and CHIMO [6][23], commonly use alpha-beta search [3][11][14][25][26], similar to computer chess. When performing alpha-beta search, it is critical [5] to measure the strength of positions accurately based on the features of pieces, locations, mobility, threat and protection, king safety, etc. Position strength is usually evaluated from the weights of designated features. In the past, these features were carefully chosen and their weights were manually tuned together with experts in most programs. However, this work becomes difficult and time-consuming when the number of features grows.

To avoid manually tuning the evaluation functions, two issues need to be taken into consideration during the design of evaluation functions: define features and automatically tune these feature weights. For the former, many proposed *n-tuple* networks, which require very little expert knowledge through its use of large numbers of features while allowing easy access. It was successfully applied to Othello [15], Connect4 [22] and 2048 [18][27]. For the latter, machine learning methods were used to tune feature weights to improve the strength of programs [2][19][20][21][24]. One of the successful methods, called *comparison training*, was employed in backgammon [19][20], shogi and chess programs [21][24]. Since Chinese chess is similar to shogi and chess, it is worth investigating whether the same technique can be applied to training for Chinese chess. In contrast, although deep reinforcement learning [16][17] recently made significant success on Go, chess and shogi, the required computing power (5000 TPUs as mentioned in [17]) is too costly for many developers.

This paper includes an n-tuple network with features that take into consideration the relationship of material combinations and positional information from individual features in Chinese chess. We then propose a novel evaluation method that incorporates the so-called *tapered eval* [8] into comparison training. Finally, we investigate batch training, which helps parallelize the process of comparison training.

Our experiments show significant improvements through the use of comparison training. With the same features, the automatically tuned comparison training feature weights achieved a win rate of 86.58% against the weights that were hand-tuned. We then improved by adding more features, most importantly n-tuple features, into the above trained version. This improved version achieved a win rate of 81.65% against the trained version without additional features.

The rest of this paper is organized as follows. Section II reviews related work on n-tuple networks and comparison training. Section III describes features used in Chinese chess programs and Section IV proposes a comparison training method. Section V presents the experimental results. Section VI makes concluding remarks.

## II. Background

In this section, we review Chinese chess in Subsection II.A, the evaluation function using n-tuple networks in Subsection II.B, the comparison training algorithm in Subsection II.C and stage-dependent features in Subsection II.D.

### A. Chinese Chess

Chinese chess is a two-player zero-sum game played on a $9 \times 10$ board. Each of two players, called *red* and *black*, has seven types of pieces: one king (`K`/`k`), two guards (`G`/`g`), two ministers (`M`/`m`), two rooks (`R`/`r`), two knights (`N`/`n`), two cannons (`C`/`c`) and five pawns (`P`/`p`). The abbreviations are uppercase for the red pieces and lowercase for the black pieces. Red plays first, then the two players take turns making one

This work was supported in part by the Ministry of Science and Technology of Taiwan under contracts MOST 106-2221-E-009-139-MY2, 106-2221-E-009-140-MY2, 106-2221-E-305-016-MY2 and 106-2221-E-305-017-MY2.

The authors[1] are with the Department of Computer Science, National Chiao Tung University, Hsinchu 30050, Taiwan. (e-mail: wenjie0723@gmail.com, icwu@csie.nctu.edu.tw (correspondent), and tinghan.wei@gmail.com)

The author[2] is with the Department of Computer Science and Information Engineering, National Taipei University, New Taipei City 23741, Taiwan. (e-mail: jcchen@mail.ntpu.edu.tw)

move at a time. The goal is to win by capturing the opponent's king. Rules are described in more detail in [28].

*B. N-tuple Network*

As mentioned in Section I, many researchers chose to use n-tuple networks since they require very little expert knowledge while allowing easy access. Evaluation functions based on n-tuple networks are linear, and can be formulated as follows.

$$\text{eval}(w, s) = w^T \varphi(s), \quad (1)$$

where $\varphi(s)$ is a feature vector that indicates features in a position $s$, and $w$ is a weight vector corresponding to these features.

In [18], an n-tuple network was defined to be composed of $m$ $n_i$-tuples, where $n_i$ is the size of the $i$-th tuple. The $n_i$-tuple is a set of $c_{i1} \times c_{i2} \times ... \times c_{in_i}$ features, each of which is indexed by $v_{i1}, v_{i2}, ..., v_{in_i}$, where $0 \le v_{ij} < c_{ij}$ for all $j$. For example, one 6-tuple covers six designated squares on the Othello board [15] and includes $3^6$ features, where each square is empty or occupied by a black or white piece. Another example is that one 4-tuple covers four designated squares on the 2048 board [27] and includes $16^4$ features, since each square has 16 different kinds of tiles.

For linear evaluation, the output is a linear summation of feature weights for all occurring features. Thus, for each tuple, since one and only one feature occurs at a time, the feature weight can be easily accessed by table lookup. If an n-tuple network includes $m$ different tuples, we need $m$ lookups.

*C. Comparison Training*

Tesauro introduced a learning paradigm called comparison training for training evaluation functions [19]. He implemented a neural-net-based backgammon program NEUROGAMMON [20], which won the gold medal in the first Computer Olympiad in 1989. He also applied comparison training to tuning a subset of the weights in Deep Blue's evaluation function [21]. For the game of shogi, Hoki [13] used comparison training to tune the evaluation function of his program BONANZA, which won the 2006 World Computer Shogi Championship. Thereafter, this algorithm was widely applied to most top shogi programs.

Comparison training is a supervised learning method. Given a training position $s$ and its *best* child position $s_1$, all the other child positions are compared with $s_1$. The best child $s_1$ is assumed to be the position reached by an expert's move. The goal of the learning method is to adjust the evaluation function so that it tends to choose the move to $s_1$. The features involved in the evaluation function for comparison are extracted for tuning. Let $w^{(t)}$ be the weight vector in the $t$-th update. An online training method is called *averaged perceptron* [9], described as follows.

$$w^{(t)} = w^{(t-1)} + d^{(t)},$$
$$d^{(t)} = \frac{1}{|S^{(t)}|} \sum_{s_i \in S^{(t)}} (\varphi(s_1) - \varphi(s_i)), \quad (2)$$

where $S^{(t)}$ is the set of child positions of $s$ whose evaluation values are higher than that of $s_1$ when $w^{(t-1)}$ is applied to the evaluation function, $|S^{(t)}|$ is the set's cardinality, and $\varphi(s_i)$ is the feature vector of $s_i$. In this paper, $d^{(t)}$ is called the *update*

*quantity* for the $t$-th update. For each iteration, all training positions are trained once, and at the end of the iteration, the weight vector $w^*$ is updated to the average of all weight vectors, $w^{(0)}$ to $w^{(T)}$, where $T$ is the total number of training positions. Then, $w^{(0)}$ is set to $w^*$ at the beginning of the next iteration. Incidentally, $w^*$ can be thought of as a measure of the training quality of one iteration by counting the number of correct moves of test positions. The whole training process stops when this number decreases.

The research in [21] observed that the feature weights can be tuned more accurately if the above evaluation values for all $s_i$ are more accurate, e.g., when the value of $s_i$ is obtained through a deeper search. Thus, $d$-*ply (comparison) training* was proposed by replacing $s_i$ in Formula (2) by the leaf $l_i$ on $s_i$'s *principle variation (PV)* in the minimax search with depth $d$, as shown in Fig. 1.

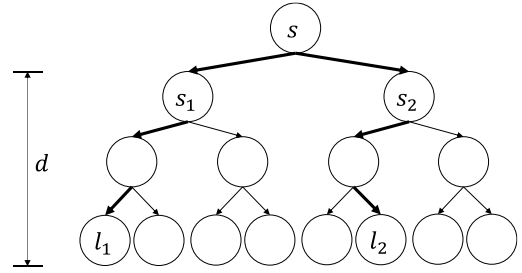

Fig. 1. An extension of comparison training ($d = 3$).

*D. Stage-dependent Features*

In many game-playing programs, the choice of features and the feature weights depended on the game stages. This is because the importance of features varies as games progress. In Othello, thirteen stages were designated according to the number of pieces on the board [4]. In 2048, three or more stages were designated according to certain tiles that occur [27]. In chess evaluation functions, tapered eval was used to calculate the stages based on the remaining pieces on the board, and implemented by most chess programs, such as FRUIT, CRAFTY and STOCKFISH.

In tapered eval, each feature has two weights representing its weights at the opening and at the endgame. A game is divided into many stages, and the weight of a feature is calculated by a linear interpolation of the two weights for each stage. More specifically, the weight vector $w$ in Formula (1) is replaced by the following.

$$w = \alpha(s)w_o + (1 - \alpha(s))w_e, \quad (3)$$

where $w_o$ is the weight vector at the opening, $w_e$ is that at the endgame, and the game stage index $\alpha(s)$, $0 \le \alpha(s) \le 1$, indicates how close it is to the opening for position $s$.

Tapered eval is also well suited for Chinese chess programs. For example, experts commonly evaluate cannons higher than knights in the opening, but less in the endgame. Hence, it is necessary to use different weights for the same feature in each stage.

III. DEFINING FEATURES FOR CHINESE CHESS

This section describes three types of features in our evaluation function for Chinese chess in Subsections III.A, III.B and III.C. Subsection III.D summarizes those features.

## A. Features for Individual Pieces

This type of features consists of the following three feature sets. The *material* (abbr. *MATL*) indicates piece types. The *location* (abbr. *LOC*) indicates the occupancies of pieces on the board. Symmetric piece locations share the same feature since the importance of symmetric locations should be the same. E.g., with symmetry, there are only 50 LOC features for knights. The *mobility* (abbr. *MOB*) indicates the number of places that pieces can be moved to without being captured. E.g., in Fig. 2, the mobility of piece **R** at location f4 is counted as seven since **R** can only move to d4, e4, f2, f3, f6, f8 and h4 safely.

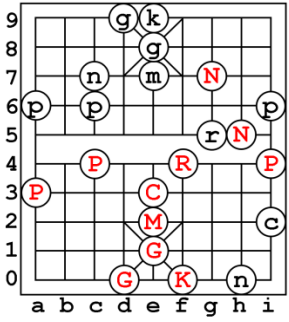

Fig. 2. A position for feature explanation.

## B. Features for King Safety

This type of features indicates the threats to the king. Two effective threats are *attacking the king's adjacency* (abbr. *AKA*) and *safe potential check* (abbr. *SPC*). When the king's adjacent locations are under attack, the mobility of the king is reduced so that it increases the chance for the opponent to checkmate. The AKA value is the summation of the number of such attacks. For example, in Fig. 2, the AKA value to **k** is four, with contributions from one **R**, two from **N**, and one from **C**.

The SPC value is the summation of the number of moves, made by opponent pieces $p$, which can check the king in the next ply while $p$ will not be captured immediately after the move. For example, in Fig. 2, the SPC value to **K** is three, where one is from **r**, one from **n**, and one from **c**. Each AKA or SPC value is given a weight individually and is seen as a feature.

## C. Features for Piece Relation

This type of features consists of three feature sets. The first feature set is *chasing the opponent's pieces* (abbr. *COP*), indicating that piece $p$ chases opponent's piece $q$ if $q$ is immediately under $p$'s attack.

The second feature set, *MATL2*, identifies the material relations between the numbers of pieces of two piece types. For example, it is well known in the Chinese chess community that it is favorable towards the opponent if a player does not have both guards and the opponent has at least one knight. Since there are at most two guards and two knights, we can use one 2-tuple of $(2 + 1) \times (2 + 1)$ features to represent the material relation of MATL2 for guards and knights. Each combination of two piece types composes a 2-tuple of MATL2.

The third feature set, *LOC2*, extracts the locational relation of two pieces, which may include attack or defense strategies. For example, in Fig. 2, both pieces **n** at c7 and **p** at c6 are not in a good shape since **p** blocks the move of **n**. Another example is that two **N**s protect each other, preventing from opponent's attack (e.g., by **r**). One 2-tuple of $50 \times 90$ can be used to represent the location relation of LOC2 for two knights when the left-right symmetry of piece locations is considered. Thus, one 2-tuple of LOC2 is used for each combination of two pieces.

## D. Summary of Features

Table I lists the numbers of features for each feature set [1].

Table I. Feature sets for Chinese chess.

| Feature set | MATL | LOC | MOB | AKA | SPC | COP | MATL2 | LOC2 |
|---|---|---|---|---|---|---|---|---|
| # of features | 6 | 194 | 26 | 5 | 5 | 32 | 462 | 119,960 |

## IV. COMPARISON TRAINING

In Subsection IV.A, we investigate comparison training for feature weights based on tapered eval. In Subsection IV.B, we present a batch training for comparison training. In Subsection IV.C, we discuss the issue for initialization.

### A. Comparison Training for Tapered Eval

For tapered eval, $w$ does not physically exist and is calculated from $w_o$ and $w_e$ for a position $s$ based on the proportion, $\alpha(s)$ and $1 - \alpha(s)$, as in Formula (3). Thus, intuitively, the update quantity for updating $w_o$ and $w_e$ should be also proportional to $\alpha(s)$ and $1 - \alpha(s)$ respectively, as follows.

$$w_o^{(t)} = w_o^{(t-1)} + \alpha(s)d^{(t)} \quad (4)$$
$$w_e^{(t)} = w_e^{(t-1)} + (1 - \alpha(s))d^{(t)} \quad (5)$$

Unfortunately, when $0 < \alpha(s) < 1$, the update quantity actually updated is less than what it should be, making the training imbalanced. For example, if $\alpha(s) = 0.5$, then

$$\begin{aligned} w^{(t)} &= 0.5 w_o^{(t)} + 0.5 w_e^{(t)} \\ &= 0.5 w_o^{(t-1)} + 0.5 w_e^{(t-1)} + 0.5 d^{(t)} \\ &= w^{(t-1)} + 0.5 d^{(t)}, \end{aligned} \quad (6)$$

where only $0.5 d^{(t)}$ is updated. Therefore, we propose new formulas to make the training balanced as follows.

$$w_o^{(t)} = w_o^{(t-1)} + \frac{\alpha(s)}{\alpha(s)^2 + (1 - \alpha(s))^2} d^{(t)} \quad (7)$$

$$w_e^{(t)} = w_e^{(t-1)} + \frac{1 - \alpha(s)}{\alpha(s)^2 + (1 - \alpha(s))^2} d^{(t)} \quad (8)$$

The update quantity is proved to be equivalent to that in Formula (2) as follows.

$$\begin{aligned} w^{(t)} &= \alpha(s) w_o^{(t)} + (1 - \alpha(s)) w_e^{(t)} \\ &= \left(\alpha(s) w_o^{(t-1)} + (1 - \alpha(s)) w_e^{(t-1)}\right) \\ &\quad + \frac{\alpha(s)^2 + (1 - \alpha(s))^2}{\alpha(s)^2 + (1 - \alpha(s))^2} d^{(t)} \\ &= w^{(t-1)} + d^{(t)} \end{aligned} \quad (9)$$

### B. Batch Training and Parallelization

In this subsection, we investigate batch training for comparison training. Batch training was commonly used to help improve the training quality of machine learning. Since batch

---

[1] For MOB, we only consider the two strong piece types, R and N. For AKA and SPC, their possible values are 1 to 4 and ≥ 5. For LOC2, the number is calculated by removing symmetry and illegal locations.

training also supports multiple threads, it helps speed up the training process.

The most time-consuming part during training is searching the training positions for the leaf nodes on PVs, such as $l_1$ and $l_2$ in Fig. 1. Once search is complete, some leaf positions are selected into $S^{(t)}$ of Formula (2) as described in Subsection II.C, and the new $w^{(t)}$ is updated based on the values of $w^{(t-1)}$.

For batch training, we update the weight vector once after searching a batch of $N$ training positions. Namely, Formula (2) is changed to the following.

$$w^{(t)} = w^{(t-N)} + \sum_{i=t-N+1}^{t} d^{(i)}, \quad (10)$$

where $d^{(i)}$ is calculated based on the weight $w^{(t-N)}$. Let $T$ be the total number of training positions. Then, only $\lceil T/N \rceil$ updates are needed in one iteration.

The above batch training provides a way for parallelism, using multiple threads to search $N$ positions in one batch. Namely, each thread grabs one position to search whenever idling. The computation of averaged weight vector $w^*$ remains unchanged.

### C. Weight Initialization

Before training, the weights of MATL are initialized as in Table II. Other feature weights are initialized to zero.

Table II. Initial weights of training for MATL.

| Guard | Minister | Rook | Knight | Cannon | Pawn |
|---|---|---|---|---|---|
| 350 | 350 | 2000 | 950 | 950 | 300 |

### V. EXPERIMENTS

In our experiments, the training data collected from [10] include 63,340 game records of expert players whose Elo ratings exceed 2000. From these game records, 1.4 million positions were randomly selected, one million for training and the rest for testing.

For benchmarking, 1071 opening positions were selected based on the frequencies played by experts, similar to [7]. Considering from the perspective of both players, a total of 2142 games were played in each experiment. Game results were judged according to the Asian rules [28], and a game with more than 400 plies was judged as a draw.

We list all the versions done in our experiments in Subsection V.A, and describe the effect of our training methods including tapered eval in Subsection V.B. The experiment and comparison of all versions are described in Subsection V.C. Subsection V.D shows experiments for batch training, and Subsection V.E for 1-, 2-, 3- and 4-ply training.

### A. List of Versions for Evaluation Functions

We used fourteen evaluation versions in experiments, based on the feature sets in Table I. These versions are listed in Table III. The version EVAL0 consists of features in the three feature sets, MATL, LOC and MOB. EVAL1 includes the feature sets used by EVAL0 plus the feature set AKA, and similarly, EVAL2 to EVAL7 include extra feature sets as shown in Table III. EVAL7 includes all the three feature sets, AKA, SPC and COP.

EVAL8 to EVAL13 contain 2-tuple feature sets, MATL2 and/or LOC2, in addition to EVAL0 and EVAL7. The number of features used in each version is also listed in the third column of the table.

Table III. Features of evaluation functions.

| Evaluation versions | Features sets | # of features |
|---|---|---|
| EVAL0 | MATL+LOC+MOB | 226 |
| EVAL1 | EVAL0+AKA | 231 |
| EVAL2 | EVAL0+SPC | 231 |
| EVAL3 | EVAL0+COP | 258 |
| EVAL4 | EVAL0+AKA+SPC | 236 |
| EVAL5 | EVAL0+AKA+COP | 263 |
| EVAL6 | EVAL0+SPC+COP | 263 |
| EVAL7 | EVAL0+AKA+SPC+COP | 268 |
| EVAL8 | EVAL0+MATL2 | 688 |
| EVAL9 | EVAL7+MATL2 | 730 |
| EVAL10 | EVAL0+LOC2 | 120,186 |
| EVAL11 | EVAL7+LOC2 | 120,228 |
| EVAL12 | EVAL0+MATL2+LOC2 | 120,648 |
| EVAL13 | EVAL7+MATL2+LOC2 | 120,690 |

### B. Training and Tapered Eval

This subsection describes the effect of our training methods including tapered eval. The comparison training method described in Section IV.A is incorporated into the Chinese chess program, CHIMO, which won the second place in Computer Olympiad 2015. In the rest of this subsection, the original version (without training) is called the *hand-tuned* version (since all the feature weights are tuned manually), while the incorporated version is called the *trained* version. For simplicity of analysis, we only consider the features in EVAL0. Each move takes 0.4 seconds on an Intel® Core™ i5-4690 processor, which is about four hundred thousand nodes per move. In the rest of this section, the time setting for each move is the same.

Table IV presents the performance comparisons of the hand-turned and trained versions, with and without tapered eval (marked with stars in the table when tapered eval is used). Note that the trained version is based on 3-ply training. From the table, the trained version clearly outperforms the hand-tuned with and without tapered eval by win rates of 86.58% and 82.94% respectively. This shows that comparison training does help improve strength significantly. Moreover, both trained and hand-tuned versions using tapered eval also outperform those without it by win rates of 62.75% and 53.76% respectively.

Table IV. Experiment results for comparison training and tapered eval. The version marked with a star uses tapered eval.

| Players | Win rate |
|---|---|
| trained vs. hand-tuned | 82.94% |
| trained* vs. hand-tuned* | 86.58% |
| hand-tuned* vs. hand-tuned | 53.76% |
| trained* vs. trained | 62.75% |

In the rest of the experiments, all versions in Table III use tapered eval and 3-ply training, unless specified explicitly.

### C. Comparison for Using Different Feature Sets

This subsection compares all the versions listed in Table III against EVAL0. Fig. 3 shows the win rates of EVAL1-EVAL13 versions playing against EVAL0.

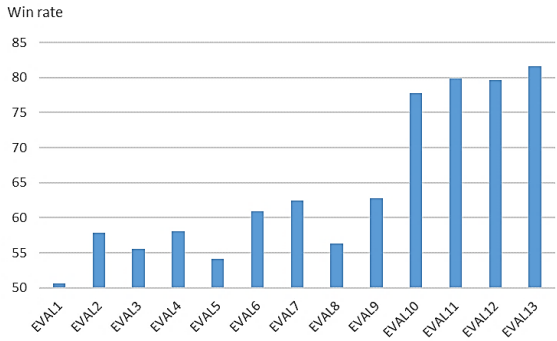

Fig. 3. Win rates of all versions against EVAL0.

In general, as in Fig. 3, the more feature sets are added, the higher the win rates are. The only exception is the case that adds AKA from EVAL3 to EVAL5. Consider the three non-tuple feature sets AKA, SPC and COP. From EVAL1-EVAL3, SPC improves more than AKA and COP. EVAL7 performs better than EVAL1-EVAL6.

Consider the two 2-tuple feature sets, MATL2 and LOC2. Clearly, all versions including LOC2, EVAL10-EVAL13, significantly outperform others. In contrast, the versions including MATL2 do not. EVAL13 that includes all feature sets reaches a win rate of 81.65%, the best among all the versions.

### D. Batch Training and Parallelization

First, we analyze the quality of batch training with three batch sizes of 50, 100 and 200 for comparison training. Fig. 4 shows the win rates of the versions trained with three batch sizes against those without batch training. The results show that the versions with batch training perform slightly better for EVAL0-EVAL9 and roughly equally for EVAL10-EVAL13. For EVAL13, the win rates are 50.56%, 50.21% and 49.53% with batch sizes 50, 100 and 200 respectively. Hence, in general, batch training does not improve playing strength significantly in this paper.

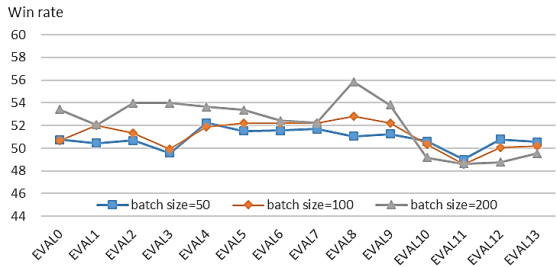

Fig. 4. Win rates of all versions with batch training against those without batch training.

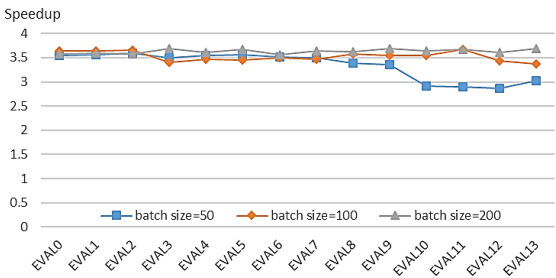

Fig. 5. Speedups.

Batch training also helps significantly on training speedup. Fig. 5 shows that for all versions EVAL0-EVAL13 the training speedups on 4-core CPUs are 2.87-3.59 times as fast as those on single-core CPUs for batch size 50, 3.37-3.68 times for 100, and 3.57-3.7 times for 200.

### E. Comparison from 1-ply to 4-ply Training

This subsection analyzes the strength of 1-, 2-, 3- and 4-ply training. From Fig. 6, all versions trained with 2-, 3- and 4-ply generally outperform those with 1-, 2- and 3-ply training respectively, except for EVAL10. Namely, for 4-ply vs. 3-ply, EVAL5 improves most by a win rate of 55.56%, while EVAL10 shows no improvement, with a win rate of 49.37%. For $d$-ply training where $d \geq 5$, we tried one case for EVAL13 (performing best among the above versions) with 5-ply training. 5-ply training in this case shows no improvement over 4-ply training, with a win rate of only 49.42%. Other experiments for $d \geq 5$ were omitted due to time constraints.

In our experiments, 3-ply training is sufficient since for EVAL10-EVAL13, 4-ply and 3-ply training performed about the same. Although for EVAL0-EVAL9, 4-ply training outperforms 3-ply training, EVAL10-EVAL13 are the stronger versions that are used when competing.

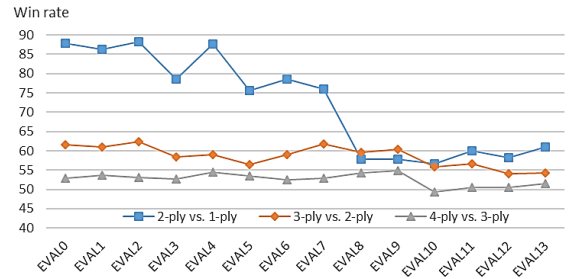

Fig. 6. Win rates of $d$-ply training, $1 \leq d \leq 4$.

### VI. CONCLUSIONS

This paper designs comparison training for computer Chinese chess. First, we propose to extract large numbers of features by leveraging n-tuple networks that require very little expert knowledge while allowing easy access. Second, we also propose a novel method to incorporate tapered eval into comparison training,

In our experiments, EVAL0 with 3-ply training outperforms our original hand-tuned version by a win rate of 86.58%. With 3-ply training, the version EVAL13, including a 2-tuple network (consisting of LOC2 and MATL2), outperforms EVAL0 by a win rate of 81.65%. Moreover, EVAL13 with 4-ply training performs better than 3-ply training by a win rate of 51.49%. The above shows efficiency and effectiveness to use comparison training to tune feature weights. In our experiments, batch training does not improve playing strength significantly. However, it speeds up the training process by a factor of 2.87-3.7 with four threads. We incorporated the above into the Chinese chess program CHIMO and won gold in Computer Olympiad 2017 [23].

Our results also show that all versions including LOC2 perform much better. This justifies that, with the help of a 2-tuple network, LOC2 is able to extract useful features with very little expert knowledge. This result also shows the potential for locational relations among three or more pieces. However, add-ing just one more piece increases the number of features dramatically. Note that the number of features for LOC2 is 119,960. This makes it difficult to complete training in a reasonable amount of time. We leave it as an open problem for further investigation.

REFERENCES

[1] Allis, L.V., *Searching for Solutions in Games and Artificial Intelligence*, Ph.D. Thesis, University of Limburg, Maastricht, The Netherlands, 1994.
[2] Baxter, J., Tridgell, A. and Weaver, L., Learning to Play Chess Using Temporal Differences, *Machine Learning* 40(3), 243-263, 2000.
[3] Beal, D.F., A Generalised Quiescence Search Algorithm, *Artificial Intelligence* 43(1), 85-98, 1990.
[4] Buro, M., Experiments with Multi-ProbCut and a New High-Quality Evaluation Function for Othello, *Technical Report* 96, NEC Research Institute, 1997.
[5] Campbell, M., A. Joseph Hoane Jr., and Hsu, F.-H., Deep Blue, *Artificial Intelligence* 134(1-2), 47-83, 2002.
[6] Chen, J.-C., Yen, S.-J., and Chen, T.-C., Shiga Wins Chinese Chess Tournament, *ICGA Journal* 36(3), 173-174, 2013.
[7] Chen, J.-C., Wu, I-C., Tseng, W.-J., Lin, B.-H., Chang, C.-H., Job-Level Alpha-Beta Search, *IEEE Transactions on Computational Intelligence and AI in Games* 7(1), 28-38, 2015.
[8] Chess Programming Wiki, Taper Eval, [Online], Available: https://chessprogramming.wikispaces.com/Tapered+Eval
[9] Collins, M., Discriminative Training Methods for Hidden Markov Models: Theory and Experiments with Perceptron Algorithms, In *EMNLP'02*, pp. 1-8, 2002.
[10] Dong Ping Xiang Qi, A Chinese chess website (in Chinese) [Online]. Available: http://www.dpxq.com/.
[11] Hsu, S.-C., Introduction to Computer Chess and Computer Chinese Chess, *Journal of Computer* 2(2), 1-8, 1990. (in Chinese)
[12] Iida, H., Sakuta, M., and Rollason, J., Computer Shogi, *Artificial Intelligence* 134, 121-144, 2002.
[13] Kaneko, T. and Hoki, K., Analysis of Evaluation-Function Learning by Comparison of Sibling Nodes, In *Advances in Computer Games 13*, LNCS 7168, 158-169, 2012.
[14] Knuth, D.E. and Moore, R.W., An Analysis of Alpha-Beta Pruning, *Artificial Intelligence* 6(4), 293-326, 1975.
[15] Lucas, S. M., Learning to Play Othello with N-tuple Systems, *Australian Journal of Intelligent Information Processing* 4, 1-20, 2007.
[16] Silver, D. *et al.*, Mastering the Game of Go with Deep Neural Networks and Tree Search, *Nature* 529, 484-489, 2016.
[17] Silver, D. *et al.*, Mastering Chess and Shogi by Self-Play with a General Reinforcement Learning Algorithm, arXiv:1712.01815, 2017
[18] Szubert, M. and Jaśkowski, W., Temporal Difference Learning of N-tuple Networks for the Game 2048, In *2014 IEEE Conference on Computational Intelligence and Games (CIG)*, pp. 1-8, 2014.
[19] Tesauro, G., Connectionist Learning of Expert Preferences by Comparison Training. *Advances in Neural Information Processing Systems* 1, 99-106, Morgan Kaufmann, 1989.
[20] Tesauro, G., Neurogammon: a Neural Network Backgammon Program, *IJCNN Proceedings* III, 33-39, 1990.
[21] Tesauro, G., Comparison Training of Chess Evaluation Functions. In: Machines that learn to play games, pp. 117-130, Nova Science Publishers, Inc., 2001.
[22] Thill , M., Koch, P. and Konen, W., Reinforcement Learning with N-tuples on the Game Connect-4, In *Proceedings of the 12th International Conference on Parallel Problem Solving from Nature - Volume Part I (PPSN'12)*, pp. 184-194, 2012.
[23] Tseng, W.-J., Chen, J.-C. and Wu, I-C., Chimo Wins Chinese Chess Tournament, *ICGA Journal*, to appear.
[24] Ura, A., Miwa, M., Tsuruoka, Y., and Chikayama, T., Comparison Training of Shogi Evaluation Functions with Self-Generated Training Positions and Moves, *CG 2013*, 2013.
[25] Ye, C. and Marsland, T.A., Selective Extensions in Game-Tree Search, Heuristic Programming in Artificial Intelligence 3, pp. 112-122. Ellis Horwood, Chichester, UK, 1992.
[26] Ye, C. and Marsland, T.A., Experiments in Forward Pruning with Limited Extensions, *ICCA Journal* 15(2), 55-66, 1992.
[27] Yeh, K.-H., Wu, I-C., Hsueh, C.-H., Chang, C.-C., Liang, C.-C. and Chiang, H., Multi-Stage Temporal Difference Learning for 2048-like Games, *IEEE Transactions on Computational Intelligence and AI in Games,* 9(4), 369-380, 2017.
[28] Yen, S.-J., Chen, J.-C., Yang, T.-N. and Hsu, S.-C., Computer Chinese Chess, *ICGA Journal* 27(1), 3-18, 2004.

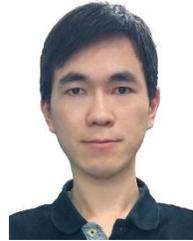

**Wen-Jie Tseng** received his B.S. in Applied Mathematics from National Chung Hsing University and M.S. in Computer Science from National Chiao Tung University in 2006 and 2008, respectively. He is currently a Ph.D. candidate in the Department of Computer Science, National Chiao Tung University. His research interests include artificial intelligence and computer games. He is the leader of the team developing the Chinese chess program, named CHIMO, which won the gold medal in Computer Olympiad 2017.

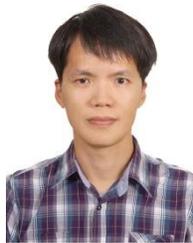

**Jr-Chang Chen** is an associate professor of the Department of Computer Science and Information Engineering at National Taipei University. He received his B.S., M.S. and Ph.D. degrees in Computer Science and Information Engineering from National Taiwan University in 1996, 1998, and 2005 respectively. He served as the Secretary General of Taiwanese Association for Artificial Intelligence in 2015 and 2016. Dr. Chen's research interests include artificial intelligence and computer games. He is the co-author of the two Chinese chess programs named ELP and CHIMO, and the Chinese dark chess program named YAHARI, which have won gold and silver medals in the Computer Olympiad tournaments.

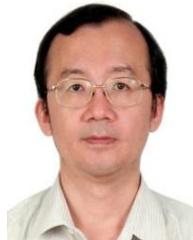

**I-Chen Wu** (M'05-SM'15) is with the Department of Computer Science, at National Chiao Tung University. He received his B.S. in Electronic Engineering from National Taiwan University (NTU), M.S. in Computer Science from NTU, and Ph.D. in Computer Science from Carnegie-Mellon University, in 1982, 1984 and 1993, respectively. He serves in the editorial boards of the IEEE Transactions on Computational Intelligence and AI in Games, ICGA Journal and Journal of Experimental & Theoretical Artificial Intelligence. He also served as the presi-dent of the Taiwanese Association of Artificial Intelligence in 2015 and 2016. His research interests include artificial intelligence, machine learning, computer games, and volunteer computing.

Dr. Wu introduced the new game, Connect6, a kind of six-in-a-row game. Since then, Connect6 has become a tour-nament item in Computer Olympiad. He led a team develop-ing various game playing programs, including CGI for Go and CHIMO for Chinese chess, winning over 30 gold medals in international tournaments, including Computer Olympiad. He wrote over 120 papers, and served as chairs and committee in over 30 academic conferences and organizations, including the conference chair of IEEE CIG conference 2015.

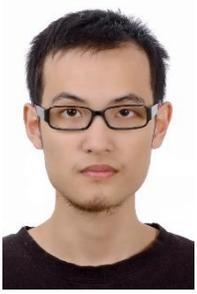

**Tinghan Wei** received his B.AS. in Electrical Engineering from University of British Columbia in 2008. He is currently a Ph.D. candidate in the Department of Computer Science, National Chiao Tung University. His research interests include artificial intelligence, machine learning and computer games.